\pdfoutput=1
\documentclass[11pt]{article}

\usepackage[final]{acl} 
\usepackage{times}
\usepackage{latexsym}
\usepackage{graphicx}
\usepackage[T1]{fontenc}
\usepackage[utf8]{inputenc}
\usepackage{amsmath}
\usepackage{changepage}
\usepackage{stix,bbding,pifont,utfsym,fontawesome}
\usepackage{adjustbox}
\usepackage{rotating}
\usepackage{array}
\usepackage{arydshln}
\usepackage{hyperref}
\usepackage{enumitem}
\usepackage{booktabs}
\usepackage{longtable}
\usepackage{caption}
\usepackage{hyperref}
\usepackage{multirow}
\usepackage{colortbl} 

\usepackage{arydshln}

\makeatletter
\def\adl@drawiv#1#2#3{%
        \hskip.5\tabcolsep
        \xleaders#3{#2.5\@tempdimb #1{1}#2.5\@tempdimb}%
                #2\z@ plus1fil minus1fil\relax
        \hskip.5\tabcolsep}
\newcommand{\cdashlinelr}[1]{%
  \noalign{\vskip\aboverulesep
           \global\let\@dashdrawstore\adl@draw
           \global\let\adl@draw\adl@drawiv}
  \cdashline{#1}
  \noalign{\global\let\adl@draw\@dashdrawstore
           \vskip\belowrulesep}}
\makeatother

\title{\textbf{\color{olive}\fontfamily{qag}\selectfont GrEmLIn}: A Repository of Green Baseline Embeddings for 87 Low-Resource Languages Injected with Multilingual Graph Knowledge}
\author{Daniil Gurgurov$^1$ \quad Rishu Kumar$^1$ \quad Simon Ostermann$^{1,2}$ \\
    $^1$German Research Center for Artificial Intelligence (DFKI) \\
    $^2$Centre for European Research in Trusted AI (CERTAIN) \\
    {\small \texttt{ \{daniil.gurgurov, rishu.kumar, simon.ostermann\}@dfki.de }}
    }

\begin{document}

\maketitle

\begin{abstract}
Contextualized embeddings based on large language models (LLMs) are available for various languages, but their coverage is often limited for lower resourced languages. Using LLMs for such languages is often difficult due to a high computational cost; not only during training, but also during inference. Static word embeddings are much more resource-efficient ("green"), and thus still provide value, particularly for very low-resource languages. There is, however, a notable lack of comprehensive repositories with such embeddings for diverse languages. To address this gap, we present \textbf{\color{olive}\fontfamily{qag}\selectfont GrEmLIn}, a centralized repository of green, static baseline embeddings for 87 mid- and low-resource languages. We compute \textbf{\color{olive}\fontfamily{qag}\selectfont GrEmLIn} embeddings with a novel method that enhances GloVe embeddings by integrating multilingual graph knowledge, which makes our static embeddings \textbf{competitive with LLM representations, while being parameter-free at inference time}. Our experiments demonstrate that \textbf{\color{olive}\fontfamily{qag}\selectfont GrEmLIn} embeddings outperform state-of-the-art contextualized embeddings from E5 on the task of lexical similarity. They remain competitive in extrinsic evaluation tasks like sentiment analysis and natural language inference, with average performance gaps of just 5-10\% or less compared to state-of-the-art models, given a sufficient vocabulary overlap with the target task, and underperform only on topic classification. Our code and embeddings are publicly available at \url{https://huggingface.co/DFKI}\footnote{All vectors are available on Huggingface as single model pages. Each page starts with \textit{DFKI/glove}.}.
\end{abstract}

\section{Introduction}

Word embedding methods have revolutionized natural language processing (NLP) by capturing semantic relationships between words using co-occurrence statistics from large text corpora \cite{mikolov2013efficient, pennington2014glove, bojanowski2017enriching}. This data-driven approach has significantly improved performance across numerous NLP tasks \cite{DBLP:journals/corr/abs-1711-00043, xie-etal-2018-neural, almeida2019word}. 

While contextual representations like the ones based on BERT \cite{devlin2018bert}, RoBERTa \cite{liu2019roberta}, and GPT \cite{radford2019language} nowadays provide better performance than static embeddings in many tasks, their training is computationally expensive \cite{strubell-etal-2019-energy, bommasani2021opportunities} and ineffective for data-scarce languages due to their data hunger and the curse of multilinguality \cite{conneau-etal-2020-unsupervised}. Some approaches of efficient adaptation of especially large language models (LLMs) to languages other than English have been investigated in recent years \cite{pfeiffer-etal-2020-mad,vykopal2024soft}. However, even such approaches still require hardware during runtime, as embeddings need to be computed based on a forward pass for each new text that is processed. This is often prohibitive in low-resource (hardware) scenarios, and inefficient in terms of energy use. Also, such approaches are often not tailored to low-resource languages.

In contrast, static word embeddings continue to play a crucial role in specific tasks such as bias detection and removal \cite{gonen2019lipstick, manzini2019black}, explaining word vector spaces \cite{vulic-etal-2020-good, bommasani-etal-2020-interpreting}, and information retrieval \cite{yan2018improving}. Static word embeddings have the advantage of being \textbf{parameter-free} at inference time, as no neural network needs to be loaded for computing such representations; just a dictionary lookup is required. This makes them both attractive for low-resource hardware scenarios, and \textbf{much more environment-friendly} \cite{strubell-etal-2019-energy, dufter-etal-2021-static}. Existing resources for multingual embedding data bases \cite{ferreira-etal-2016-jointly, bojanowski2017enriching, grave-etal-2018-learning} often suffer from limited scope and outdated data, potentially worsening their ability to capture the dynamic nature of language and adequately support low-resource languages. We want to fill this gap by providing \textbf{\color{olive}\fontfamily{qag}\selectfont GrEmLIn}, a large database of static word embeddings for 87 mid- and low-resource languages.

As for LLMs, the training of word embeddings suffers from the lack of high-quality data in low-resource languages (to a smaller degree). Incorporating other types of data for improving word representations is thus beneficial especially for low-resource languages. Knowledge graphs provide such an alternative to textual knowledge, with rich semantic and multilingual sources of information, including synonyms, antonyms, morphological forms, definitions, etimological relations, translations, and more \cite{miller1995wordnet, navigli2012babelnet, speer2017conceptnet}. Such structured and cross-lingual information can be used to improve the quality of classical word representations \cite{faruqui2014retrofitting, sakketou2020constrained}, which are only trained on co-occurence statistics. 

To that end, we propose a new simple yet effective method for including graph information into word embeddings based on \citet{mikolov2013exploiting}. We learn a projection matrix to map static embeddings to a combined space, effectively overcoming the limitations of retrofitting approaches that only enhance a limited vocabulary. This method combines the strengths of traditional word embeddings with the structured, multilingual information from knowledge graphs, resulting in more accurate and informative representations. 

In summary, our contributions in this work are two-fold: First, we present \textbf{\color{olive}\fontfamily{qag}\selectfont GrEmLIn}, a centralized resource of static word embeddings for 87 mid- and low-resource languages, specifically focusing on word embeddings trained with GloVe \cite{pennington2014glove}. Second, we propose an effective method to improve embeddings by incorporating more knowledge in the form of multilingual knowledge graphs, which is especially important for low-resource languages, where resources are usually very scarce. Our code is publicly available on GitHub\footnote{\url{https://github.com/d-gurgurov/GrEmLIn-Green-Embeddings-LRLs}}. 

\section{Related Work}
We briefly describe the most prominent graph knowledge sources, word embeddings, and existing methods for improving embeddings with graphs.

\paragraph{Graph knowledge sources.}
Among most used knowledge graphs for natural language are WordNet \cite{miller1995wordnet} and BabelNet \cite{navigli2012babelnet}. WordNet is a lexical database that organizes English words into sets of synonyms called synsets, providing short definitions and usage examples. BabelNet is a multilingual encyclopedic dictionary and semantic network, which integrates lexicographic and encyclopedic knowledge from WordNet, Wikipedia, etc., focused on named entities. 
In our work, we use ConceptNet \cite{speer2017conceptnet}, a multilingual, domain-general knowledge graph that connects words and phrases from various natural languages with labeled, weighted edges representing relationships between terms. Unlike other knowledge graphs, ConceptNet is not a monolingual collection of named entities but focuses on commonly used words and phrases across multiple languages.

\paragraph{Word embeddings.}
\textit{Word2Vec} \cite{mikolov2013efficient} uses shallow neural networks to produce word vectors. It comes in two types: Continuous Bag of Words (CBOW) and Skip-gram. CBOW predicts a word given its context, while Skip-gram predicts the context given a word. 
\textit{GloVe} (Global Vectors for Word Representation) \cite{pennington2014glove} word embeddings are created by aggregating global word-word co-occurrence statistics from a corpus. The resulting vectors capture both local and global semantic relationships. 
\textit{FastText} \cite{bojanowski2017enriching} extends Word2Vec by representing words as bags of character n-grams, capturing subword information and handling out-of-vocabulary words more effectively. FastText is particularly useful for morphologically rich languages. 
\textit{Numberbatch}, part of the ConceptNet project \cite{speer2017conceptnet}, is a set of word embeddings that integrates knowledge from ConceptNet with distributional semantics from GloVe and Word2Vec. Numberbatch uses a retrofitting approach \cite{faruqui2014retrofitting} to enhance embeddings with structured semantic knowledge. Retrofitting often results in a limited vocabulary for underrepresented languages \cite{speer2017conceptnetsemeval} since the retrofitting process relies on existing semantic relationships in ConceptNet to adjust the original embeddings.

\paragraph{Improving Embeddings with Knowledge Graphs. }
There are various methods to improve word embeddings by incorporating external knowledge graphs or semantic networks \cite{dieudonat2020exploring}. \textit{Retrofitting} \cite{faruqui2014retrofitting} is a post-processing technique that adjusts pre-trained word embeddings using information from knowledge graphs or semantic lexicons. The key idea is to infer new vectors that are close to their original embeddings while also being close to their neighbors in the graph or lexicon. This is achieved by minimizing an objective function that balances the distance between the new vectors and the original embeddings, as well as the distance between connected nodes. 
\textit{Expanded retrofitting} \cite{speer2017conceptnet}, used for ConceptNet Numberbatch, optimizes over a larger vocabulary including terms from the knowledge graph not present in the original embeddings, but it still does not retrofit all the words in the original embedding space. Other existing methods that integrate contextualized embeddings with knowledge graph embeddings often use attention mechanisms, as demonstrated by works such as \citeauthor{peters2019knowledge} (\citeyear{peters2019knowledge}), \citeauthor{sun2019ernie} (\citeyear{sun2019ernie}), and \citet{gurgurov-etal-2024-adapting}. These methods specifically enhance BERT embeddings by incorporating external knowledge bases. 


\section{Method}
We propose a method for merging GloVe embeddings with graph-based embeddings derived from ConceptNet knowledge, while preserving the vocabulary size of GloVe, following two steps: 
First, we use singular value decomposition (SVD) \cite{eckart1936approximation} on concatenated word embeddings from GloVe and pointwise mutual information (PMI) based graph embeddings \cite{speer2017conceptnet} to generate a shared embedding space. We do so for the part of the vocabulary that is shared between GloVe and the knowledge graph. Second, we learn a linear transformation from GloVe into this joined space to obtain embeddings for all words in the original GloVe vocabulary. 

\subsection{GloVe Embeddings}
We train GloVe embeddings using the original code. The model is trained by stochastically sampling nonzero elements from the co-occurrence matrix over 100 iterations, to produce 300-dimensional vectors. We use a context window of 10 words to the left and 10 words to the right. Words with fewer than 5 co-occurrences are excluded for languages with over 1 million tokens in the training data, and the threshold is set to 2 for languages with smaller datasets. We use data from CC100\footnote{\url{https://huggingface.co/datasets/cc100}} \cite{wenzek-etal-2020-ccnet, conneau-etal-2020-unsupervised} for training the static word embeddings. We set \(x_{max} = 100\), \(\alpha = \frac{3}{4}\), and use AdaGrad optimization \cite{duchi2011adaptive} with an initial learning rate of 0.05.

\subsection{Graph Embeddings}
To build ConceptNet-based word embeddings, we follow the method used for constructing ConceptNet Numberbatch embeddings \cite{speer2017conceptnet}. We represent the ConceptNet graph as a sparse, symmetric term-term matrix, where each cell is the sum of the occurences of all edges connecting the two terms. Unlike the original method, we do not discard terms connected to fewer than three edges, as we deal with low-resource languages.

We calculate embeddings from this matrix by applying pointwise mutual information (PMI) with context distributional smoothing of 0.75, clipping negative values to yield positive PMI (PPMI), which follows practical recommendations by \cite{levy-etal-2015-improving}. We then reduce the dimensionality to 300 using truncated SVD and combine terms and contexts symmetrically to form a single matrix of word embeddings, called ConceptNet-PPMI. This matrix captures the overall graph structure of ConceptNet. 

We compute ConceptNet-PPMI embeddings for the entire ConceptNet, covering 304 languages, which we call \textit{PPMI (All)}. Further, we construct separate graph embedding spaces, \textit{PPMI (Single)}, for each language, using only the portion of ConceptNet for that language. This approach is adopted since the initial co-occurence matrices for individual languages are less sparse while still being multilingual in nature.

\subsection{Singular Value Decomposition (SVD)}
We first concatenate GloVe and PPMI vectors for all words that are in the shared vocabulary, resulting in 600-dimensional vectors\footnote{PPMI embeddings are normalized to be in the range of the Glove embeddings}. Afterwards, we reduce the dimensionality and remove some of the variance coming from redundant features. The matrix \(M\) representing merged GloVe and ConceptNet-PPMI can be approximated with a truncated SVD:
\[ M \approx U \Sigma V^T \]
where \(\Sigma\) is truncated to a \( k' \times k' \) diagonal matrix of the \( k' \) largest singular values, and \( U \) and \( V \) are correspondingly truncated to have only these \( k' \) columns. \( U \) is then used as a matrix mapping the original vocabulary to a smaller set of features\footnote{We dismiss the weighting of \( U \) by the singular values from \(\Sigma\), which was noted to work better for semantic tasks \cite{levy-etal-2015-improving}}.

\subsection{Linear Transformation}
To obtain embeddings for the entire vocabulary from the original GloVe embedding space (i.e. not only the common words), we find a linear projection matrix between the spaces and project the GloVe embeddings onto the merged embedding space, similar to \citet{Mikolov2013ExploitingSA}, using a gradient descent optimization on a linear regression model.

Given a set of word pairs and their associated vector representations \(\{x_i, z_i\}_{i=1}^n\), where \( x_i \in \mathbb{R}^{d_1} \) is the GloVe representation of word \( i \), and \( z_i \in \mathbb{R}^{d_2} \) is the PPMI representation from ConceptNet, our goal is to find a transformation matrix \( W \) such that \( W x_i \) approximates \( z_i \). 

\( W \) can be learned by solving the following optimization problem:
\[ \min_W \sum_{i=1}^n \| W x_i - z_i \|^2 \]
which we solve as a linear regression problem with stochastic gradient descent optimization. 

The resulting projection matrix is used to project the GloVe embeddings onto the merged embedding space.

\section{Experiments}
In this section, we describe the selected languages, tasks, and experiments conducted to evaluate the effectiveness of our proposed method. 

\subsection{Languages}
We trained GloVe embeddings for 87 languages from the CC100 dataset \cite{wenzek-etal-2020-ccnet}, focusing on languages categorized as low-resource (class 0 to 3) based on \citeauthor{joshi-etal-2020-state}'s classification (\citeyear{joshi-etal-2020-state}). For 72 of these languages, present in both CC100 and ConceptNet, we generated additional graph embeddings. The merging process involves enhancing the original GloVe embeddings with graph knowledge via SVD-reduced PPMI integration. 
Further details about these languages, including common vocabulary size between GloVe and ConceptNet, can be found in Part \ref{appendix:common} of the Appendix.

\subsection{Evaluation Data}
We assess the embeddings using both intrinsic and extrinsic evaluation tasks. The intrinsic evaluation is performed using the MultiSimLex dataset \cite{vulic2020multi}, which provides manually annotated data on semantic similarity consisting of 1888 examples across 12 languages, 4 of which overlap with our work. This task focuses on measuring the strength of similarity between word pairs (e.g., "lion – cat") independently of relatedness, making it a good test for how well embeddings capture semantic similarity. 

For extrinsic evaluations, we focus on three downstream NLP tasks: Sentiment Analysis (SA), Topic Classification (TC), and Natural Language Inference (NLI). Due to the limited availability of intrinsic datasets for most low-resource languages, we prioritize these tasks to reflect real-world use cases, where high-quality word embeddings are crucial.

For SA, we compile data for 23 languages from multiple open sources, prioritizing mid- and low-resource languages for broader coverage across typological families. The details of these data sources are listed in Table \ref{tab:sa-sources} in the Appendix. Some datasets, such as those for Swahili, Nepali, Uyghur, Latvian, Slovak, Slovenian, Uzbek, Bulgarian, Yoruba, Bengali, Hebrew, and Telugu, are highly imbalanced in terms of class distribution. To mitigate this, we apply random undersampling to create a balanced version of the datasets. This step allows for a more robust comparison of the embeddings' performance in low-data settings.

We further evaluate the embeddings on the TC task using the SIB-200 dataset \cite{adelani-etal-2024-sib}. It offers multilingual data for topic classification, covering 200 languages, and was specifically designed to improve natural language understanding for under-resourced languages. Our experiments cover 57 languages, chosen based on their availability in both ConceptNet and CC100. The task is framed as a multi-label classification with the data distributed along 7 different classes. The dataset provides predefined train, validation and test splits, which consist of 701, 99, and 204 examples, respectively.

Lastly, we evaluate the embeddings on the NLI task, using the XNLI dataset \cite{conneau-etal-2018-xnli}. The XNLI dataset provides multilingual NLI examples for 15 languages, and for our experiments, we selected 5 of these languages: Swahili, Urdu, Greek, Thai, and Bulgarian. These languages were selected based on the availability of our GloVe, PPMI-enhanced embeddings, and the NLI dataset. Evaluating embeddings on the NLI task tests their ability to understand logical relationships between sentence pairs, an important capability for higher-level NLU tasks. Due to the simplicity of our models, we only utilize validation and test splits, consisting of 2,490 and 5,010 examples, respectively, for training and testing, excluding the original training split of nearly 400,000 examples.

\subsection{Experimental Setup}
We evaluate the embeddings using a Support Vector Machine (SVM) classifier \cite{boser1992training} for all extrinsic tasks—SA, TC, and NLI—reporting macro-averaged F1 scores for fair comparison. For the intrinsic MultiSimLex task, we use Spearman's Rank Correlation \cite{spearman1961proof} to assess how well the embeddings' similarity predictions align with human annotations.

For extrinsic tasks, sentence representations are constructed by summing word embeddings, which is a standard approach in NLP \cite{mikolov2013distributedrepresentationswordsphrases, bowman-etal-2015-large, williams-etal-2018-broad}, and then used as input features for the SVM. The SVM model is trained with a Radial Basis Function (RBF) kernel, which is commonly used for nonlinear classification problems. The regularization parameter \( C \) is fixed at 1 for GloVe-based embeddings, balancing the trade-off between maximizing the margin and minimizing classification errors. This setup minimizes the impact of hyperparameters on the resulting scores.

For the NLI task, sentence representation follows the same method as above, but with an added step. We concatenate the sentence embeddings of the two input sentences (premise and hypothesis) to form the final input representation for the SVM. This approach enables the model to capture the relationship between the two sentences.

As for baselines, we use three strong pre-trained models:
\begin{itemize}
    \item FastText \cite{grave2018learningwordvectors157}, a word embedding model that extends the traditional skip-gram model by representing words as bags of character n-grams, allowing it to effectively handle out-of-vocabulary words.
    \item XLM-R-base \cite{conneau-etal-2020-unsupervised}, a transformer-based multilingual model. We obtain sentence embeddings by summing the model's last hidden states.
    \item E-5-base \cite{wang2024multilinguale5textembeddings}, a state-of-the-art multilingual sentence embedding model known for its strong performance in multilingual tasks. This serves as a high-quality benchmark for our comparisons.
\end{itemize}

\begin{table*}[h]
\centering
\small
    \def\arraystretch{1.05}
    \begin{tabular}{l!{\vrule width 0.5pt}lcc!{\vrule width 0.5pt}cccc}
        \toprule
        \multirow{2}{*}{\rotatebox[origin=c]{90}{Cov.}}&\multirow{2}{*}{ISO} & \multicolumn{2}{c}{Contextualized} & \multicolumn{4}{c}{Static} \\
        \cmidrule(lr){3-4} \cmidrule(lr){5-8}
         && E-5-B & X-B & FT & G & GP(S) & GP(A) \\
        \midrule
    \multirow{3}{*}{\rotatebox[origin=c]{90}{>90\%}}&et   & \textbf{.19}  & .03   & .447  & .341  & \textbf{.452}  & .422  \\ 
    &he   & \textbf{.218} & .057  & .426  & .336  & \textbf{.436}  & .429  \\
    \cdashlinelr{2-8}
    &\textbf{Avg}.   & \textbf{.204} & .044  & .437  & .339  & \textbf{.444}  & .426 \\
    \midrule
    \multirow{3}{*}{\rotatebox[origin=c]{90}{<90\%}}&cy   & \textbf{.112} & .039  & .346  & .276  & \textbf{.366}  & .357  \\ 
    &sw   & \textbf{.212} & .011  & \textbf{.408} & .24   & .319   & .324  \\
    \cdashlinelr{2-8}
    &\textbf{Avg}.   & \textbf{.162} & .003  & \textbf{.377}  & .258  & .343  & .341 \\
    \midrule
    &\textbf{All avg.}   & \textbf{.183} & .034 & \textbf{.407}  & .298  & .393  &  .383 \\
    \bottomrule
    \end{tabular}
\caption{Spearman's correlation scores on MultiSimLex across 4 languages, for E-5-B, XLM-R-B, FastText, GloVe (G), GloVe + PPMI (GP), Single and All, sorted by GloVe vocabulary coverage. The horizontal solid line separates languages with over 90\% coverage (above) from those with less (below). \textbf{Bold} numbers indicate the maximum per line, for static and contextualized.}
\label{tab:results_simlex}
\end{table*}

For the XLM-R-base and E-5 embeddings, we adjust the regularization parameter \( C \) to 100. This adjustment accounts for the higher dimensionality of these embeddings, as lower \( C \) values constrain their performance.

\section{Results}

We distinguish between static and contextualized embeddings by first comparing the static embeddings against each other, and then comparing them to the contextualized ones. The results from E-5-B are provided for reference but cannot be directly compared to our static embeddings due to the reasons outlined in Section \ref{sec:context}.

\subsection{Semantic Similarity} We evaluate the performance of the embeddings on the lexical semantics task using the MultiSimLex dataset, focusing on 4 languages: Estonian, Welsh, Swahili, and Hebrew.

As shown in Table \ref{tab:results_simlex}, the GloVe+PPMI (Single) embeddings achieve the highest correlation scores for 3 out of 4 languages, demonstrating their ability to capture semantic similarities. For Swahili, FastText achieves the best result, although GloVe+PPMI remains competitive. In contrast, contextual embeddings such as XLM-R-base struggle in this intrinsic evaluation task, achieving lower correlation scores across all languages, which supports \citet{vulic2020multi}. E-5 performs better than XLM-R but does not surpass the best-performing static embeddings.

These results underscore the continued relevance of graph-enhanced static embeddings in lexical semantic tasks, particularly for low-resource languages where training data may be scarce.

\subsection{Sentiment Analysis} 
We evaluate the performance of the proposed GloVe+PPMI embeddings on the SA task for 23 mid- and low-resource languages. Table \ref{tab:results_sa} presents the results for this task. Our findings show that both GloVe+PPMI (Single) and GloVe+PPMI (All) embeddings consistently outperform the original GloVe embeddings across most languages. GloVe+PPMI (Single) improves performance for 19 out of 23 languages, while GloVe+PPMI (All) improves results for 18 out of 23 languages when compared to GloVe. 

When comparing GloVe with FastText embeddings, we observe that GloVe outperforms FastText in 12 out of 23 languages, with some languages showing comparable results. 

\begin{table}[h]
\small
\def\arraystretch{1.05}

\resizebox{\columnwidth}{!}{
\begin{tabular}{l!{\vrule width 1.0pt}lc:c!{\vrule width 0.5pt}cccc}
        \toprule
        \multirow{2}{*}{\rotatebox[origin=c]{90}{Cov.}}&\multirow{2}{*}{ISO} & \multicolumn{2}{c}{Contextualized} & \multicolumn{4}{c}{Static} \\
        \cmidrule(lr){3-4} \cmidrule(lr){5-8}
        & & E-5-B & X-B & FT & G & GP(S) & GP(A) \\
        \midrule
    \multirow{4}{*}{\rotatebox[origin=c]{90}{>90\%}}& ka   & \textbf{.9}    & .845   & .855    & .861         & \textbf{.87}  & .861    \\
    & sl   & \textbf{.881}  & .832 & .743    & .749         & .779         & \textbf{.788} \\
    & ro   & \textbf{.926}  & .872 & .803    & .805         & \textbf{.85}  & .847    \\
    \cdashlinelr{2-8}
    & \textbf{Avg.}   & \textbf{.902} & .85  & .8  & .805  & \textbf{.833}  & .832 \\
    \midrule
    \multirow{11}{*}{\rotatebox[origin=c]{90}{>80\%}} & he   & \textbf{.929}  & .811   & .782    & .788         & \textbf{.824} & .822    \\
    & si   & \textbf{.895}  & .831   & .846    & .848         & .85          & \textbf{.857} \\
    & sw   & \textbf{.773}  & .665   & .697    & .68          & .701         & \textbf{.714} \\
    & ug   & \textbf{.881}  & .61    & .792    & .746         & \textbf{.811} & \textbf{.811} \\
    & lv   & \textbf{.801}  & .74    & .749    & .783         & \textbf{.787} & \textbf{.787} \\
    & te   & \textbf{.854}  & .831 & .798    & .806         & .808         & \textbf{.817} \\
    & sk   & \textbf{.911}  & .854 & .73     & .756         & \textbf{.806} & .805    \\
    & mr   & \textbf{.912}  & .886   & .888    & .903         & \textbf{.905} & .902    \\
    & bg   & \textbf{.884}  & .721   & .793    & .786         & .801         & \textbf{.805} \\
    & mk   & \textbf{.817}  & .736 & .682    & \textbf{.716} & .711         & .7      \\
    \cdashlinelr{2-8}
    & \textbf{Avg.}   & \textbf{.866} & .769  & .776 & .781  & .8  & \textbf{.802} \\
    \midrule
    \multirow{11}{*}{\rotatebox[origin=c]{90}{<80\%}} & su   & \textbf{.855}  & .829 & .805    & .798         & \textbf{.822} & .812    \\
    & am   & \textbf{.861}  & .782   & .815    & \textbf{.881} & .86          & .88     \\
    & ne   & \textbf{.666}  & .519   & .666    & .643         & .674         & \textbf{.688} \\
    & da   & \textbf{.972}  & .927 & .895    & .863         & \textbf{.908} & .903    \\
    & uz   & \textbf{.858}  & .807   & \textbf{.822} & .808         & .806         & .806    \\
    & bn   & \textbf{.938}  & .837   & \textbf{.889} & .875         & \textbf{.881} & .878    \\
    & ur   & \textbf{.818}  & .757 & .678    & .676         & \textbf{.746} & .745    \\
    & az   & \textbf{.787}  & .762 & \textbf{.75}  & .744         & .746         & .745    \\
    & cy   & \textbf{.834}  & .795   & .798    & .77          & .789         & \textbf{.801} \\
    & yo   & \textbf{.764}  & .634   & .696    & .721         & .709         & \textbf{.738} \\
    \cdashlinelr{2-8}
    & \textbf{Avg. }  & \textbf{.835} & .765  & .781  & .778  & .794  & \textbf{.8} \\
    \midrule
    & \textbf{All avg.}   & \textbf{.857}  & .778 & .781    & .783         & .802 & \textbf{.805}    \\
    \bottomrule
\end{tabular}
}
\caption{Macro Average F1 Scores for Sentiment Analysis per language, sorted by GloVe vocabulary coverage. Horizontal solid lines indicate 90\% and 80\% coverage by GloVe. \textbf{Bold} numbers indicate the maximum per line, for static and contextualized.}
\label{tab:results_sa}
\end{table}

In contrast, XLM-R-base performs better than all static embedding configurations for 9 out of 23 languages, and E-5 outperforms most static embedding variants. While this underscores the power of contextualized models, the enhanced GloVe+PPMI embeddings remain competitive, with a drop of only 5\% in performance, especially in low-resource settings. This suggests that static embeddings, when enriched with multilingual graph knowledge, remain competitive and provide a lightweight and efficient zero-parameter alternative for resource-constrained environments.

\begin{table}[ht]
    \def\arraystretch{1.05}
    \small
    \resizebox{\columnwidth}{!}{
    \begin{tabular}{l!{\vrule width 0.5pt}lcc!{\vrule width 0.5pt}cccc}
        \toprule
        \multirow{2}{*}{\rotatebox[origin=c]{90}{Cov.}} & \multirow{2}{*}{ISO} & \multicolumn{2}{c}{Contextualized} & \multicolumn{4}{c}{Static} \\
        \cmidrule(lr){3-4} \cmidrule(lr){5-8}
        & & E-5-B & X-B & FT & G & GP(S) & GP(A) \\
        \midrule
        \multirow{4}{*}{\rotatebox[origin=c]{90}{>80\%}}& bg & \textbf{.563} & .465 & .465 & .441 & \textbf{.481} & .477 \\
        & el & \textbf{.546} & .455 & .484 & .456 & \textbf{.496} & .488 \\
        & sw & \textbf{.539} & .437 & .471 & .438 & .466 & \textbf{.468} \\
        \cdashlinelr{2-8}
        & \textbf{Avg.}   & \textbf{.549} & .452  & .473  & .445  & \textbf{.481}  & .478 \\
        \midrule
        \multirow{3}{*}{\rotatebox[origin=c]{90}{<80\%}} & ur & \textbf{.540} & .472 & .412 & .44 & \textbf{.473} & .471 \\
        & th & \textbf{.538} & .461 & .275 & .284 & .292 & \textbf{.3} \\
        \cdashlinelr{2-8}
        & \textbf{Avg.}   & \textbf{.539} & .467  & .344  & .362  & .383  & \textbf{.386} \\
        \midrule
        & \textbf{All avg.} & \textbf{.545} & .458 & .421 & .412 & \textbf{.442} & .441 \\
        \bottomrule
    \end{tabular}
    }
    \caption{Macro Average F1 Scores for Natural Language Inference per language, sorted by GloVe vocabulary coverage. The horizontal solid line separates languages with over 80\% coverage (above) from those with less (below). \textbf{Bold} numbers indicate the maximum per line, for static and contextualized. }
    \label{tab:results_xnli}
\end{table}

\subsection{Natural Language Inference} 
In the NLI task using the XNLI dataset, we again observe consistent improvements in performance with the enhanced GloVe embeddings (Table \ref{tab:results_xnli}). While GloVe outperforms FastText for only 2 out of 5 languages, the use of PPMI (Single) and PPMI (All) results in better performance for all 5 languages. 

In comparison, XLM-R performs better than the static embedding variants for 1 out of the 5 languages, and E-5 outperforms all models in all languages. While transformer models like XLM-R excel in capturing complex semantic relationships between sentences, the performance of GloVe+PPMI remains competitive, with a drop of only 6\% given a sufficient vocabulary overlap, especially in improving sentence-level reasoning and inference capabilities in low-resource languages. 

\subsection{Topic Classification} 
The results of the topic classfication task using the SIB-200 dataset are the only results where the contextualized models seem to have a clear advantage. The drop for a vocabulary coverage of over 95\% is only at 10\%, but over all languages the drop averages at 20\% when comparing the best contextualized with the best static model (Table \ref{tab:results_sib}). GloVe embeddings outperform FastText for 27 out of 57 languages, using GloVe+PPMI (Single) boosts performance for 37 out of 57 languages, and GloVe+PPMI (All) enhances performance for 48 out of 57 languages. 

XLM-R gives better performance than all static embedding configurations for only 24 out of 57 languages, but E-5 performs better than all static embeddings, showcasing some strengths of contextualized embeddings in multilingual tasks.

\begin{table}[pt!]
    \def\arraystretch{1.05}
    \small
    \resizebox{\columnwidth}{!}{
    \begin{tabular}{l!{\vrule width 1.0pt}lcc!{\vrule width 1.0pt}cccc}
        \toprule
        \multirow{2}{*}{\rotatebox[origin=c]{90}{Cov.}} & \multirow{2}{*}{ISO} & \multicolumn{2}{c}{Contextualized} & \multicolumn{4}{c}{Static} \\
        \cmidrule(lr){3-4} \cmidrule(lr){5-8}
        & & E-5-B & X-B & FT & G & GP(S) & GP(A) \\
        \midrule
        \multirow{14}{*}{\rotatebox[origin=c]{90}{>95\%}}& ro & \textbf{.891} & .707 & .405 & .561 & .686 & \textbf{.704} \\
        & sk & \textbf{.866} & .707 & .522 & .67 & .667 & \textbf{.725} \\
        & bg & \textbf{.869} & .751 & .447 & .645 & .711 & \textbf{.723} \\
        & el & \textbf{.872} & .66 & .387 & .531 & \textbf{.712} & .702 \\
        & lt & \textbf{.861} & .704 & .534 & .713 & .775 & \textbf{.797} \\
        & uk & \textbf{.904} & .717 & .542 & .682 & .722 & \textbf{.745} \\
        & lv & \textbf{.858} & .709 & .608 & .737 & .732 & \textbf{.742} \\
        & sl & \textbf{.848} & .705 & .544 & .628 & .715 & \textbf{.734} \\
        & gl & \textbf{.865} & .723 & .522 & .53 & .663 & \textbf{.699} \\
        & da & \textbf{.864} & .724 & .423 & .446 & \textbf{.743} & .717 \\
        & he & \textbf{.824} & .701 & .67 & .759 & .739 & \textbf{.784} \\
        & mk & \textbf{.855} & .779 & .598 & .611 & .694 & \textbf{.719} \\
        & ms & \textbf{.846} & .748 & .634 & .694 & .738 & \textbf{.769} \\
        \cdashlinelr{2-8}
        & \textbf{Avg.}   & \textbf{.863} & .718  & .526  & .631  & .715  & \textbf{.735} \\
        \midrule
        \multirow{13}{*}{\rotatebox[origin=c]{90}{>90\%}}& et & \textbf{.823} & .655 & .583 & .589 & .572 & \textbf{.605} \\
        & be & \textbf{.836} & .698 & \textbf{.674} & .58 & .597 & .621 \\
        & az & \textbf{.853} & .742 & .667 & .698 & .668 & \textbf{.711} \\
        & eo & \textbf{.844} & .665 & .57 & .504 & .567 & \textbf{.588} \\
        & hy & \textbf{.809} & .622 & .411 & .551 & .609 & \textbf{.644} \\
        & kk & \textbf{.838} & .69 & .64 & .664 & .647 & \textbf{.69} \\
        & is & \textbf{.798} & .651 & .442 & .423 & .49 & \textbf{.534} \\
        & ka & \textbf{.763} & .694 & .591 & .684 & \textbf{.689} & \textbf{.689} \\
        & ur & \textbf{.804} & .648 & .451 & .42 & .627 & \textbf{.643} \\
        & cy & \textbf{.704} & .638 & .615 & .564 & .608 & \textbf{.694} \\
        & af & \textbf{.865} & .721 & .573 & .454 & .56 & \textbf{.59} \\
        & si & \textbf{.809} & .682 & .647 & .678 & .613 & \textbf{.695} \\
        \cdashlinelr{2-8}
        & \textbf{Avg.}   & \textbf{.812} & .676  & .572  & .567  & .604  & \textbf{.642} \\
        \midrule
        \multirow{15}{*}{\rotatebox[origin=c]{90}{>80\%}}& tl & \textbf{.85} & .624 & .656 & .65 & .707 & \textbf{.709} \\
        & bn & \textbf{.814} & .595 & .567 & .604 & .617 & \textbf{.681} \\
        & ga & \textbf{.704} & .441 & \textbf{.585} & .411 & .532 & .547 \\
        & mr & \textbf{.838} & .64 & .567 & .627 & .608 & \textbf{.676} \\
        & ky & \textbf{.783} & .665 & \textbf{.633} & .593 & .571 & .59 \\
        & gu & \textbf{.79} & .613 & \textbf{.663} & .544 & .589 & .631 \\
        & ml & \textbf{.812} & .664 & .641 & \textbf{.651} & .574 & .608 \\
        & pa & \textbf{.807} & .556 & \textbf{.566} & .474 & .521 & .565 \\
        & kn & \textbf{.803} & .604 & \textbf{.672} & .652 & .581 & .658 \\
        & ne & \textbf{.796} & .699 & .553 & .542 & .563 & \textbf{.605} \\
        & ha & \textbf{.708} & .449 & --- & .421 & .489 & \textbf{.546} \\
        & ja & \textbf{.802} & .608 & \textbf{.656} & .511 & .523 & .541 \\
        & ug & \textbf{.723} & .606 & \textbf{.642} & .556 & .583 & .622 \\
        & am & \textbf{.781} & .559 & \textbf{.585} & .515 & .472 & .555 \\
        \cdashlinelr{2-8}
        & \textbf{Avg.}   & \textbf{.787} & .595  & \textbf{.614}  & .554  & .566  & .61 \\
        \midrule
        \multirow{15}{*}{\rotatebox[origin=c]{90}{<80\%}}& su & \textbf{.765} & .561 & \textbf{.572} & .467 & .446 & .526 \\
        & so & \textbf{.642} & .388 & .442 & .363 & .403 & \textbf{.459} \\
        & ps & \textbf{.73} & .542 & \textbf{.532} & .351 & .431 & .493 \\
        & ht & \textbf{.717} & .318 & \textbf{.531} & .392 & .496 & .523 \\
        & yi & \textbf{.538} & .36 & \textbf{.532} & .341 & .384 & .453 \\
        & gd & \textbf{.54} & .341 & .404 & .23 & .397 & \textbf{.418} \\
        & xh & \textbf{.641} & .324 & --- & {.388} & .32 & .341 \\
        & yo & \textbf{.663} & .185 & \textbf{.341} & .199 & .211 & .264 \\
        & sa & \textbf{.762} & .542 & \textbf{.452} & .206 & .261 & .251 \\
        & qu & \textbf{.561} & .245 & \textbf{.294} & .175 & .167 & .153 \\
        & my & \textbf{.791} & .564 & .171 & \textbf{.228} & .207 & .163 \\
        & km & \textbf{.74} & .631 & .114 & \textbf{.125} & .117 & .109 \\
        & ku & \textbf{.657} & .202 & .09 & \textbf{.11} & .098 & .095 \\
        & lo & \textbf{.743} & .704 & --- & .183 & \textbf{.185} & .18 \\
        & wo & \textbf{.594} & .238 & --- & .058 & \textbf{.139} & .122 \\
        \cdashlinelr{2-8}
        & \textbf{Avg.}  & \textbf{.672} & .41  & \textbf{.373}  & .254  & .284  & .303 \\
        \midrule
        & \textbf{All avg.} & \textbf{.779} & .591 & .529 & .497 & .535 & \textbf{.566} \\
        \bottomrule
    \end{tabular}
    }
    \caption{Macro Average F1 Scores for Topic Classification per language, sorted by GloVe vocabulary coverage. The horizontal solid lines indicate 95\%, 90\%, and 80\% coverage by GloVe. \textbf{Bold} numbers indicate the maximum per line, for static and contextualized.}
    \label{tab:results_sib}
\end{table}

\subsection{Additional Experiment: Graph-enhanced GloVe Improvement}
To explain the improvements from injecting graph knowledge into static embeddings, we hypothesize that the size of the common vocabulary between GloVe and PPMI spaces contributes to performance gains: a larger vocabulary may lead to a better linear transformation fit, resulting in more precise projections. We investigate the relationship between common vocabulary size and embedding improvements by calculating Pearson \cite{cohen2009pearson} and Spearman \cite{spearman1961proof} correlations across all tasks (SID-200, SA, XNLI, SimLex). 

Table \ref{tab:cor} shows the correlation coefficients for each task and embedding configuration. For SID-200, the GloVe+PPMI (Single) embeddings have a Spearman correlation of 0.364, indicating a moderate monotonic relationship between common vocabulary count and improvement. However, the Pearson correlation of 0.096 suggests a weak linear relationship. A similar pattern can be observed in the other tasks, where Spearman correlations are generally higher than Pearson, highlighting the non-linear nature of the relationship. In contrast, tasks like SimLex show high correlations in both metrics, especially in the Single configuration, with Pearson and Spearman scores of 0.879 and 0.8, respectively.

When comparing results for Single and All configurations, the Single configurations tend to show stronger correlations. The All configurations have higher vocabulary overlaps between the embedding spaces due to contributions from various languages (\ref{appendix:common} of the Appendix). This is because Single configurations focus on one language, whereas All configurations include words from multiple languages, which dilutes the strength of the relationship between vocabulary overlap and performance improvement and may suggest that the Single embedding spaces provide a better representation of graph knowledge when working in a monolingual setting.

These results suggest that while improvement scores moderately depend on vocabulary coverage, the relationship isn’t strictly linear. This implies that while a larger common vocabulary can enhance performance, other factors such as graph-based semantic knowledge may play a more significant role. Figure \ref{fig:cor} in the Appendix visualizes the correlations for SA and SID-200 across all models.

\section{Discussion: Contextual vs. Static Embeddings} \label{sec:context} 
While being far behind on the tested intrinsic task, the sentence embeddings extracted from E-5, a state-of-the-art multilingual sentence embedding model, consistently outperform many other configurations across all languages on the extrinsic tasks. This superior performance can be attributed to E-5's ability to generate context-aware, sentence-level representations that capture nuanced meanings, which static embeddings, like GloVe or FastText, struggle to achieve. Unlike static word embeddings that sum individual word vectors, E-5 learns richer representations by incorporating contextual information across languages.

However, direct comparisons between E-5 and static word embeddings overlook key differences in design and use cases. E-5 is extensively trained on multilingual corpora and excels in tasks requiring complex, context-sensitive representations. In contrast, static embeddings, though simpler, are a valid alternative in low-resource or efficiency-critical scenarios, as they are effectively parameter-free during inference time. The coverage of task-specific data plays a crucial role: GloVe embeddings performe well in sentiment analysis due to broader language coverage, but poorer results in topic classification are partly linked to lower coverage in some languages. \textbf{Static embeddings remain competetive across most tested extrinsic tasks and most languages, given a good vocabulary coverage.}

Static embeddings enriched with external knowledge sources, such as graph-based information, provide significant advantages, especially in resource-limited applications where computational costs are critical. Computationally lightweight word vectors are invaluable in settings where models like E-5 are prohibitively expensive to deploy \cite{strubell-etal-2019-energy, bommasani2021opportunities}. Static embeddings also perform competitively in simpler tasks that do not heavily rely on contextual understanding \cite{dufter-etal-2021-static}, making them ideal for large-scale or real-time applications \cite{gupta-jaggi-2021-obtaining}.
Additionally, static embeddings offer a level of transparency often lacking in complex models \cite{vulic-etal-2020-good, bommasani-etal-2020-interpreting}. \textbf{Their word-level semantic relationships are easy to interpret, making them useful in applications such as bias detection or model auditing.}

Furthermore, \citeauthor{dufter-etal-2021-static} (\citeyear{dufter-etal-2021-static}) demonstrated that FastText outperformed BERT on a modified LAMA task \cite{petroni2019languagemodelsknowledgebases} across ten languages while generating just 0.3\% of BERT's carbon footprint \cite{strubell-etal-2019-energy, dufter-etal-2021-static}, despite their simplicity. This highlights the overlooked value of static embeddings when evaluating resource-intensive models, \textbf{rendering them useful as "green" baselines that are environmentally highly efficient}. 

\begin{table}[tb] 
\centering 
\small 
\begin{tabular}{lcc} 
\toprule 
Task & P & S \\ 
\midrule 
SID-200 (Single) & 0.096 & 0.364 \\ 
SID-200 (All) & 0.284 & 0.115 \\ 
SA (Single) & 0.116 & 0.261 \\ 
SA (All) & 0.254 & 0.186 \\ 
XNLI (Single) & 0.075 & 0.205 \\ 
XNLI (All) & 0.054 & 0.300 \\ 
SimLex (Single) & 0.879 & 0.800 \\ 
SimLex (All) & 0.399 & 0.105 \\ 
\bottomrule 
\end{tabular} 
\caption{Pearson and Spearman Correlations between Common Vocabulary Count and Improvement Scores} 
\label{tab:cor} \end{table}

\section{Conclusion}
In this work, we developed \textbf{\color{olive}\fontfamily{qag}\selectfont GrEmLIn}, a centralized repository of graph-enhanced GloVe embeddings for 87 mid- and low-resource languages, addressing the need for high-quality word embeddings in underrepresented languages. By merging GloVe with graph-based knowledge from ConceptNet, we enhanced the semantic richness of embeddings, leading to improved performance across tasks like semantic similarity, sentiment analysis, topic classification, and natural language inference.

Our results show that graph-enhanced GloVe outperforms the original GloVe, FastText, and even contextualized embeddings from XLM-R, offering a lightweight and environmentally efficient alternative to transformer-based models. Static embeddings have been recognized as "green" baselines, offering competitive performance at a fraction of computational cost of LLMs. This makes them ideal for low-resource settings where both computational efficiency and sustainability are key.

\section*{Limitations}
While our contribution provides baseline and graph-enhanced GloVe models for many languages, several limitations exist. First, the quality and availability of training data, particularly for low-resource languages, remain key challenges. Despite leveraging large corpora like CC100 and ConceptNet, data diversity and coverage are still limited.

Second, while our method of merging GloVe embeddings with graph-based knowledge has yielded promising results, there is room for further refinement. Future work could explore more advanced fusion and projection techniques to enhance representations for low-resource languages.

Lastly, static embeddings, even with graph enhancements, cannot fully capture contextual nuances compared to transformer-based models, which may limit their performance on tasks requiring deep contextual understanding. Balancing simplicity and efficiency with improved performance remains an ongoing challenge.

\section*{Acknowledgments}
We thank the anonymous reviewers for their helpful feedback. This work was supported by DisAI – Improving scientific excellence and creativity in combating disinformation with artificial intelligence and language technologies, a Horizon Europe-funded project under GA No. 101079164, and by the German Ministry of Education and Research (BMBF) as part of the project TRAILS (01IW24005).

\bibliography{custom}

\appendix

\onecolumn
\section*{Appendix}

\begin{table*}[htb]
\section{Language Details}
\small
\centering
\resizebox{\columnwidth}{!}{
\begin{tabular}[b]{clcccclcccc}
\toprule
\textbf{ISO code} & \textbf{Language} & \textbf{Size} & \textbf{Class} & \textbf{ConceptNet} & \textbf{ISO code} & \textbf{Language} & \textbf{Size} & \textbf{Class} & \textbf{ConceptNet}\\
\cmidrule(r){1-5}\cmidrule(l){6-10}
{\bf ss} & Swati & 86K & 1 & ✘ & {\bf sc} & Sardinian & 143K & 1 & ✓ \\
{\bf yo} & Yoruba & 1.1M & 2 & ✓ & {\bf gn} & Guarani & 1.5M & 1 & ✓ \\
{\bf qu} & Quechua & 1.5M & 1 & ✓ & {\bf ns} & Northern Sotho & 1.8M & 1 & ✘ \\
{\bf li} & Limburgish & 2.2M & 1 & ✓ & {\bf ln} & Lingala & 2.3M & 1 & ✓ \\
{\bf wo} & Wolof & 3.6M & 2 & ✓ & {\bf zu} & Zulu & 4.3M & 2 & ✓ \\
{\bf rm} & Romansh & 4.8M & 1 & ✓ & {\bf ig} & Igbo & 6.6M & 1 & ✘ \\
{\bf lg} & Ganda & 7.3M & 1 & ✘ & {\bf as} & Assamese & 7.6M & 1 & ✘ \\
{\bf tn} & Tswana & 8.0M & 2 & ✘ & {\bf ht} & Haitian & 9.1M & 2 & ✓ \\
{\bf om} & Oromo & 11M & 1 & ✘ & {\bf su} & Sundanese & 15M & 1 & ✓ \\
{\bf bs} & Bosnian & 18M & 3 & ✘ & {\bf br} & Breton & 21M & 1 & ✓ \\
{\bf gd} & Scottish Gaelic & 22M & 1 & ✓ & {\bf xh} & Xhosa & 25M & 2 & ✓ \\
{\bf mg} & Malagasy & 29M & 1 & ✓ & {\bf jv} & Javanese & 37M & 1 & ✓ \\
{\bf fy} & Frisian & 38M & 0 & ✓ & {\bf sa} & Sanskrit & 44M & 2 & ✓ \\
{\bf my} & Burmese & 46M & 1 & ✓ & {\bf ug} & Uyghur & 46M & 1 & ✓ \\
{\bf yi} & Yiddish & 51M & 1 & ✓ & {\bf or} & Oriya & 56M & 1 & ✓ \\
{\bf ha} & Hausa & 61M & 2 & ✓ & {\bf la} & Lao & 63M & 2 & ✓ \\
{\bf sd} & Sindhi & 67M & 1 & ✓ & {\bf ta\_rom} & Tamil Romanized & 68M & 3 & ✘ \\
{\bf so} & Somali & 78M & 1 & ✓ & {\bf te\_rom} & Telugu Romanized & 79M & 1 & ✘ \\
{\bf ku} & Kurdish & 90M & 0 & ✓ & {\bf pu/pa} & Punjabi & 90M & 2 & ✓ \\
{\bf ps} & Pashto & 107M & 1 & ✓ & {\bf ga} & Irish & 108M & 2 & ✓ \\
{\bf am} & Amharic & 133M & 2 & ✓ & {\bf ur\_rom} & Urdu Romanized & 141M & 3 & ✘ \\
{\bf km} & Khmer & 153M & 1 & ✓ & {\bf uz} & Uzbek & 155M & 3 & ✓ \\
{\bf bn\_rom} & Bengali Romanized & 164M & 3 & ✘ & {\bf ky} & Kyrgyz & 173M & 3 & ✓ \\
{\bf my\_zaw} & Burmese (Zawgyi) & 178M & 1 & ✘ & {\bf cy} & Welsh & 179M & 1 & ✓ \\
{\bf gu} & Gujarati & 242M & 1 & ✓ & {\bf eo} & Esperanto & 250M & 1 & ✓ \\
{\bf af} & Afrikaans & 305M & 3 & ✓ & {\bf sw} & Swahili & 332M & 2 & ✓ \\
{\bf mr} & Marathi & 334M & 2 & ✓ & {\bf kn} & Kannada & 360M & 1 & ✓ \\
{\bf ne} & Nepali & 393M & 1 & ✓ & {\bf mn} & Mongolian & 397M & 1 & ✓ \\
{\bf si} & Sinhala & 452M & 0 & ✓ & {\bf te} & Telugu & 536M & 1 & ✓ \\
{\bf la} & Latin & 609M & 3 & ✓ & {\bf be} & Belarussian & 692M & 3 & ✓ \\
{\bf tl} & Tagalog & 701M & 3 & ✘ & {\bf mk} & Macedonian & 706M & 1 & ✓ \\
{\bf gl} & Galician & 708M & 3 & ✓ & {\bf hy} & Armenian & 776M & 1 & ✓ \\
{\bf is} & Icelandic & 779M & 2 & ✓ & {\bf ml} & Malayalam & 831M & 1 & ✓ \\
{\bf bn} & Bengali & 860M & 3 & ✓ & {\bf ur} & Urdu & 884M & 3 & ✓ \\
{\bf kk} & Kazakh & 889M & 3 & ✓ & {\bf ka} & Georgian & 1.1G & 3 & ✓ \\
{\bf az} & Azerbaijani & 1.3G & 1 & ✓ & {\bf sq} & Albanian & 1.3G & 1 & ✓ \\
{\bf ta} & Tamil & 1.3G & 3 & ✓ & {\bf et} & Estonian & 1.7G & 3 & ✓ \\
{\bf lv} & Latvian & 2.1G & 3 & ✓ & {\bf ms} & Malay & 2.1G & 3 & ✓ \\
{\bf sl} & Slovenian & 2.8G & 3 & ✓ & {\bf lt} & Lithuanian & 3.4G & 3 & ✓ \\
{\bf he} & Hebrew & 6.1G & 3 & ✓ & {\bf sk} & Slovak & 6.1G & 3 & ✓ \\
{\bf el} & Greek & 7.4G & 3 & ✓ & {\bf th} & Thai & 8.7G & 3 & ✓ \\
{\bf bg} & Bulgarian & 9.3G & 3 & ✓ & {\bf da} & Danish & 12G & 3 & ✓ \\
{\bf uk} & Ukrainian & 14G & 3 & ✓ & {\bf ro} & Romanian & 16G & 3 & ✓ \\
{\bf id} & Indonesian & 36G & 3 & ✘ &  &  &  &  &  \\
\bottomrule
\end{tabular}
}
\caption{\centering Details of the reproduced CC-100 corpus available on HuggingFace, including languages with their ISO codes, data set sizes, low-resource classifications, and language availability in the ConceptNet knowledge graph.}
\label{tab:datastats}
\end{table*}

\onecolumn
\begin{table*}[htb]
\section{SA Data Details}
\begin{adjustwidth}{-1cm}{}
    \centering
    \small
    \begin{tabular}{l c | c | c c | c c c}
        \toprule
        \textbf{Language} & \textbf{ISO code} & \textbf{Source} & \textbf{\#pos} & \textbf{\#neg} & \textbf{\#train} & \textbf{\#val} & \textbf{\#test} \\
        \midrule
        Sundanese & su & \citeauthor{winata2022nusax}, \citeyear{winata2022nusax} & 378 & 383 & 381 & 76 & 304 \\
        Amharic & am & \citeauthor{tesfa2024aspect}, \citeyear{tesfa2024aspect} & 487 & 526 & 709 & 152 & 152 \\
        Swahili & sw & \citeauthor{Muhammad2023AfriSentiAT}, \citeyear{Muhammad2023AfriSentiAT}; \citeauthor{muhammad2023semeval}, \citeyear{muhammad2023semeval} & 908 & 319 & 738 & 185 & 304 \\
        Georgian & ka & \citeauthor{stefanovitch-etal-2022-resources}, \citeyear{stefanovitch-etal-2022-resources} & 765 & 765 & 1080 & 120 & 330 \\
        Nepali & ne & \citeauthor{9381292}, \citeyear{9381292} & 680 & 1019 & 1189 & 255 & 255 \\
        Uyghur & ug & \citeauthor{li2022senti}, \citeyear{li2022senti} & 2450 & 353 & 1962 & 311 & 530 \\
        Latvian & lv & \citeauthor{SprogisRikters2020BalticHLT}, \citeyear{SprogisRikters2020BalticHLT} & 1796 & 1380 & 2408 & 268 & 500 \\
        Slovak & sk & \citeauthor{pecar-etal-2019-improving}, \citeyear{pecar-etal-2019-improving} & 4393 & 731 & 3560 & 522 & 1042 \\
        Sinhala & si & \citeauthor{ranathunga2021sentiment}, \citeyear{ranathunga2021sentiment} & 2487 & 2516 & 3502 & 750 & 751 \\
        Slovenian & sl & \citeauthor{buvcar2018annotated}, \citeyear{buvcar2018annotated} & 1665 & 3337 & 3501 & 750 & 751 \\
        Uzbek & uz & \citeauthor{KuriyozovMAG19}, \citeyear{KuriyozovMAG19} & 3042 & 1634 & 3273 & 701 & 702 \\
        Bulgarian & bg & \citeauthor{martinez2021evaluating}, \citeyear{martinez2021evaluating} & 6652 & 1271 & 5412 & 838 & 1673 \\
        Yoruba & yo & \citeauthor{Muhammad2023AfriSentiAT}, \citeyear{Muhammad2023AfriSentiAT}; \citeauthor{muhammad2023semeval}, \citeyear{muhammad2023semeval} & 6344 & 3296 & 5414 & 1327 & 2899 \\
        Urdu & ur & \citeauthor{maas-EtAl:2011:ACL-HLT2011}, \citeyear{maas-EtAl:2011:ACL-HLT2011}; \citeauthor{khan2017harnessing}, \citeyear{khan2017harnessing}; \citeauthor{khan2020usc}, \citeyear{khan2020usc} & 5562 & 5417 & 7356 & 1812 & 1812 \\
        Macedonian & mk & \citeauthor{jovanoski-etal-2015-sentiment}, \citeyear{jovanoski-etal-2015-sentiment} & 3041 & 5184 & 6557 & 729 & 939 \\
        Danish & da & \citeauthor{isbister-etal-2021-stop}, \citeyear{isbister-etal-2021-stop} & 5000 & 5000 & 7000 & 1500 & 1500 \\
        Marathi & mr & \citeauthor{pingle2023l3cube}, \citeyear{pingle2023l3cube} & 5000 & 5000 & 8000 & 1000 & 1000 \\
        Bengali & bn & \citeauthor{sazzed-2020-cross}, \citeyear{sazzed-2020-cross} & 8500 & 3307 & 8264 & 1771 & 1772 \\
        Hebrew & he & \citeauthor{amram-etal-2018-representations}, \citeyear{amram-etal-2018-representations} & 8497 & 3911 & 8932 & 993 & 2483 \\
        Romanian & ro & \citeauthor{tache-etal-2021-clustering}, \citeyear{tache-etal-2021-clustering} & 7500 & 7500 & 10800 & 1200 & 3000 \\
        Telugu & te & \citeauthor{marreddy2022resource}, \citeyear{marreddy2022resource}; \citeauthor{marreddy2022multi}, \citeyear{marreddy2022multi} & 9488 & 6746 & 11386 & 1634 & 3214 \\
        Welsh & cy & \citeauthor{espinosa2021english}, \citeyear{espinosa2021english} & 12500 & 12500 & 17500 & 3750 & 3750 \\
        Azerbaijani & az & \citeauthor{azerbaijanisent}, \citeyear{azerbaijanisent} & 14000 & 14000 & 19600 & 4200 & 4200  \\
        \bottomrule
    \end{tabular}
    \caption{ \centering Sentiment Analysis Data Details}
    \label{tab:sa-sources}
\end{adjustwidth}
\end{table*}

\onecolumn
\begin{table}[htb]
\newpage
\section{Common Vocabulary Counts}
\label{appendix:common}
    \centering
    \small
    \begin{tabular}{lcc}
        \toprule
        \textbf{ISO code} & \textbf{GloVe and PPMI (Single)} & \textbf{GloVe and PPMI (All)} \\
        \midrule
        af   & 9,177   & 85,270   \\
        am   & 1,105   & 14,217   \\
        az   & 7,215   & 80,761   \\
        be   & 7,623   & 73,750   \\
        bn   & 3,962   & 38,221   \\
        bg   & 92,436  & 368,232  \\
        ku   & 3,762   & 32,499   \\
        cy   & 7,774   & 57,522   \\
        da   & 38,095  & 450,290  \\
        el   & 19,710  & 197,647  \\
        eo   & 59,476  & 161,634  \\
        et   & 14,815  & 163,666  \\
        gd   & 6,415   & 24,430   \\
        ga   & 13,871  & 65,169   \\
        gl   & 29,654  & 215,868  \\
        gu   & 3,198   & 24,575   \\
        ht   & 1,557   & 13,304   \\
        ha   & 671     & 33,824   \\
        he   & 16,032  & 153,731  \\
        hy   & 14,951  & 60,756   \\
        is   & 27,007  & 143,567  \\
        ja   & 2,607   & 41,471   \\
        kn   & 2,181   & 24,783   \\
        ka   & 17,869  & 96,066   \\
        kk   & 8,292   & 64,494   \\
        km   & 2,654   & 34,014   \\
        ky   & 2,234   & 29,915   \\
        lo   & 269,010 & 373,012  \\
        lt   & 12,485  & 200,404  \\
        lv   & 17,450  & 183,088  \\
        ml   & 4,092   & 38,864   \\
        mr   & 3,211   & 33,552   \\
        mk   & 21,692  & 93,121   \\
        my   & 3,189   & 24,319   \\
        ne   & 2,650   & 21,479   \\
        pa   & 2,282   & 16,068   \\
        ps   & 847     & 15,904   \\
        ro   & 25,704  & 366,809  \\
        sa   & 3,336   & 12,101   \\
        si   & 943     & 27,536   \\
        sk   & 14,694  & 268,576  \\
        sl   & 45,153  & 229,429  \\
        so   & 533     & 18,088   \\
        su   & 1,236   & 26,068   \\
        sw   & 6,425   & 59,906   \\
        ta   & 4,596   & 60,906   \\
        tl   & 12,563  & 42,653   \\
        ug   & 764     & 4,798    \\
        uk   & 16,397  & 327,563  \\
        ur   & 4,662   & 44,530   \\
        uz   & 3,229   & 37,704   \\
        xh   & 1,650   & 15,709   \\
        yi   & 5,177   & 18,572   \\
        ms   & 34,022  & 152,500  \\
        yo   & 558     & 5,254    \\
        qu   & 2,056   & 11,046   \\
        wo   & 999     & 18,509   \\
        th   & 45,975  & 238,502  \\
        \bottomrule
    \end{tabular}
    \caption{Common Vocabulary between GloVe and PPMI Embedding Spaces}
\end{table}

\onecolumn
\begin{table}[htb]
\newpage
\section{Vocabulary Coverage}
\centering
\small

\begin{tabular}{lcccccccc}
\toprule
\textbf{ISO code} & \multicolumn{2}{c}{\textbf{SA}} & \multicolumn{2}{c}{\textbf{SIB}} & \multicolumn{2}{c}{\textbf{XNLI}} & \multicolumn{2}{c}{\textbf{MultiSimLex}} \\
\midrule
              & \textbf{G (\%)} & \textbf{F (\%)} & \textbf{G (\%)} & \textbf{F (\%)} & \textbf{G (\%)} & \textbf{F (\%)} & \textbf{G (\%)} & \textbf{F (\%)} \\
\midrule
am            & 78.22              & 99.48           & 84.36           & 99.73           & --              & --              & --              & --              \\
su            & 78.66              & 99.92           & 79.39           & 99.94           & --              & --              & --              & --              \\
sw            & 88.24              & 100.00          & 91.32           & 99.98           & 83.68           & 99.98           &    73.94            & 100.00              \\
si            & 89.18              & 99.99           & 91.63           & 99.97           & --              & --              & --              & --              \\
ka            & 97.19              & 99.99           & 94.71           & 100.00          & --              & --              & --              & --              \\
ne            & 77.91              & 99.82           & 84.93           & 99.99           & --              & --              & --              & --              \\
ug            & 88.28              & 99.92           & 82.87           & 99.96           & --              & --              & --              & --              \\
yo            & 22.37              & 99.18           & 46.50           & 99.73           & --              & --              & --              & --              \\
ur            & 62.54              & 99.72           & 92.97           & 99.95           & 73.42           & 99.86           & --              & --              \\
mk            & 82.90              & 99.92           & 95.84           & 99.99           & --              & --              & --              & --              \\
mr            & 84.06              & 99.94           & 87.13           & 99.99           & --              & --              & --              & --              \\
bn            & 66.55              & 99.75           & 89.58           & 100.00          & --              & --              & --              & --              \\
te            & 85.66              & 99.99           & --              & --              & --              & --              & --              & --              \\
uz            & 71.17              & 99.94           & 83.61           & 99.99           & --              & --              & --              & --              \\
az            & 60.60              & 100.00          & 94.03           & 100.00          & --              & --              & --              & --              \\
bg            & 84.18              & 99.91           & 98.16           & 100.00          & 96.47           & 99.98           & --              & --              \\
sl            & 91.79              & 100.00          & 97.92           & 100.00          & --              & --              & --              & --              \\
lv            & 87.04              & 99.41           & 97.43           & 99.97           & --              & --              & --              & --              \\
sk            & 84.74              & 99.75           & 98.29           & 99.99           & --              & --              & --              & --              \\
ro            & 90.16              & 99.94           & 98.71           & 100.00          & --              & --              & --              & --              \\
he            & 89.72              & 99.74           & 97.57           & 100.00          & --              & --              & 91.79              & 100.00              \\
cy            & 51.87              & 99.91           & 90.76           & 99.98           & --              & --              &     82.73           &  100.00             \\
da            & 75.48              & 99.71           & 96.76           & 100.00          & --              & --              & --              & --              \\
el            & --              & --              & 98.15           & 99.94           & 97.34           & 100.00          & --              & --              \\
th            & --              & --              & --              & --              & 22.29           & 100.00          & --              & --              \\
af            & --              & --              & 90.05           & 99.95           & --              & --              & --              & --              \\
be            & --              & --              & 94.59           & 99.95           & --              & --              & --              & --              \\
eo            & --              & --              & 93.83           & 100.00          & --              & --              & --              & --              \\
et            & --              & --              & 94.50           & 100.00          & --              & --              &   94.70            &  99.99              \\
gd            & --              & --              & 70.48           & 99.85           & --              & --              & --              & --              \\
ga            & --              & --              & 89.97           & 99.93           & --              & --              & --              & --              \\
gl            & --              & --              & 97.33           & 99.98           & --              & --              & --              & --              \\
gu            & --              & --              & 87.11           & 99.97           & --              & --              & --              & --              \\
ht            & --              & --              & 75.89           & 99.74           & --              & --              & --              & --              \\
ha            & --              & --              & 87.20           & --              & --              & --              & --              & --              \\
hy            & --              & --              & 92.70           & 99.92           & --              & --              & --              & --              \\
is            & --              & --              & 92.22           & 99.94           & --              & --              & --              & --              \\
ja            & --              & --              & 82.97           & 99.98           & --              & --              & --              & --              \\
kn            & --              & --              & 86.82           & 100.00          & --              & --              & --              & --              \\
kk            & --              & --              & 93.11           & 100.00          & --              & --              & --              & --              \\
km            & --              & --              & 24.06           & 99.92           & --              & --              & --              & --              \\
ky            & --              & --              & 87.29           & 100.00          & --              & --              & --              & --              \\
lo            & --              & --              & 19.17           & --              & --              & --              & --              & --              \\
lt            & --              & --              & 97.70           & 100.00          & --              & --              & --              & --              \\
ml            & --              & --              & 85.13           & 100.00          & --              & --              & --              & --              \\
my            & --              & --              & 31.18           & 99.96           & --              & --              & --              & --              \\
pa            & --              & --              & 85.15           & 99.98           & --              & --              & --              & --              \\
ps            & --              & --              & 78.72           & 99.91           & --              & --              & --              & --              \\
sa            & --              & --              & 46.87           & 99.94           & --              & --              & --              & --              \\
so            & --              & --              & 79.48           & 99.89           & --              & --              & --              & --              \\
tl            & --              & --              & 89.07           & 100.00          & --              & --              & --              & --              \\
uk            & --              & --              & 97.72           & 100.00          & --              & --              & --              & --              \\
xh            & --              & --              & 62.39           & --              & --              & --              & --              & --              \\
yi            & --              & --              & 73.63           & 99.87           & --              & --              & --              & --              \\
ms            & --              & --              & 95.71           & 100.00          & --              & --              & --              & --              \\
qu            & --              & --              & 36.33           & 99.96           & --              & --              & --              & --              \\
wo            & --              & --              & 54.05           & --              & --              & --              & --              & --              \\
\bottomrule
\end{tabular}
\caption{Vocabulary Coverage by GloVe FastText Embeddings for 4 Evaluation Tasks - Sentiment Analysis, Topic Classification, Natural Language Inference, and MultiSimLex}

\end{table}

\onecolumn
\section{Correlation Between Improvement Scores and Vocabulary Overlap}
\label{appendix:cor}

\begin{figure*}[htb]
    \centering
    \includegraphics[width=\textwidth]{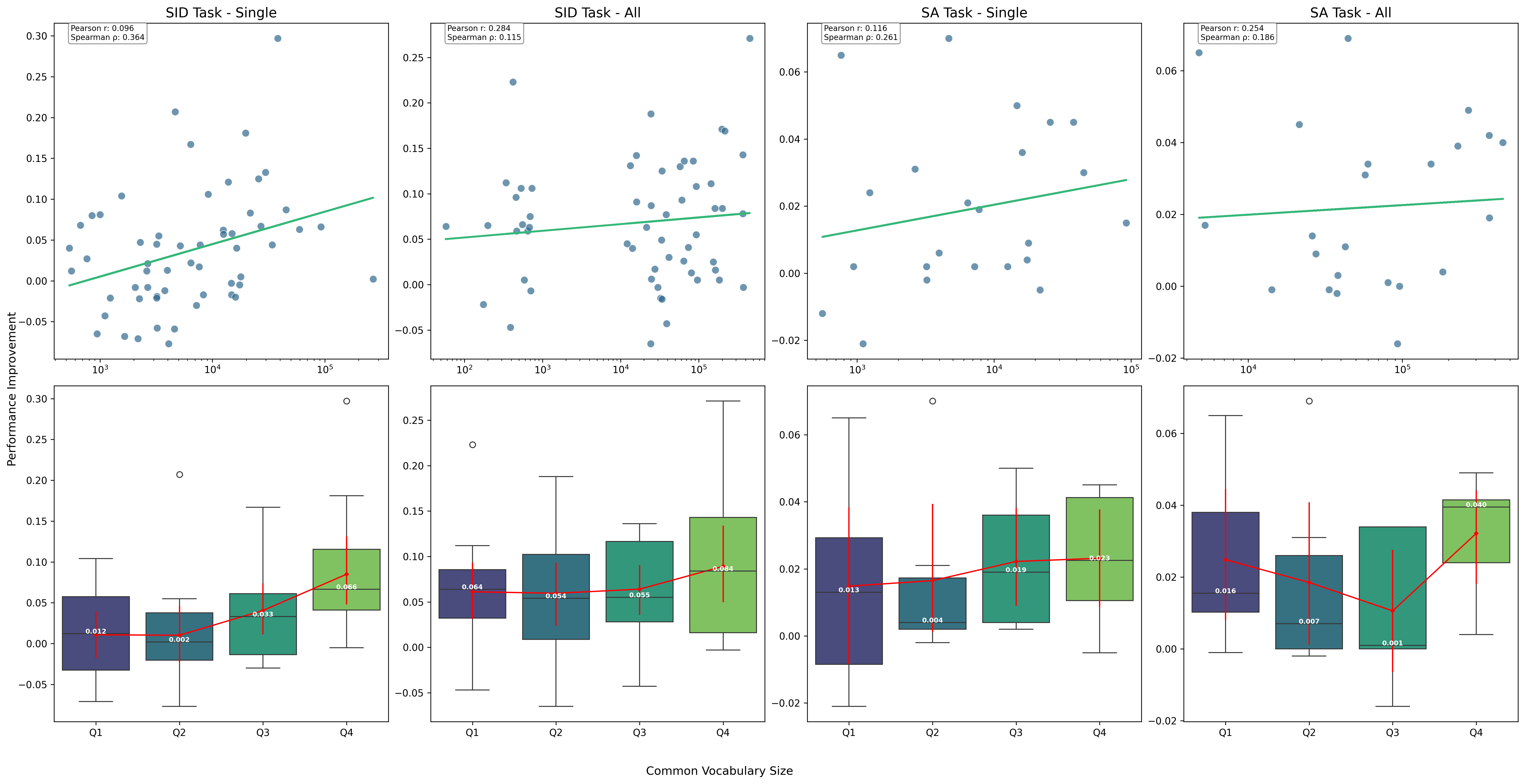}
    \caption{Scatter plots illustrating the relationship between vocabulary overlap and performance improvements across various language tasks using GloVe and graph-enhanced embeddings (G+P). Each plot shows the improvement in performance (G+P \texttt{-} GloVe) versus the common vocabulary size (log-scaled). Solid lines represent the best-fit log-linear trend.}
    \label{fig:cor}
\end{figure*}

\end{document}